\documentclass[preprint,12pt]{elsarticle}

\usepackage{xcolor}%
\usepackage{textcomp}%
\usepackage{graphicx} % Required for inserting images
\usepackage{setspace}
\usepackage{lineno}
\usepackage{comment}
\usepackage{amssymb}
\usepackage{multirow} 
\usepackage{subfigure}

\usepackage{amsthm}

\title{Intelligent Agricultural Management Considering N2O Emission and Climate Variability with Uncertainties}

\author[inst1]{Zhaoan Wang}

\author[inst1]{Shaoping Xiao}

\author[inst2]{Jun Wang}

\author[inst3]{Ashwin Parab}

\author[inst3]{Shivam Patel}

\affiliation[inst1]{organization={Department of Mechanical Engineering, Iowa Technology Institute, University of Iowa},
            addressline={3131 Seamans Center}, 
            city={Iowa City},
            postcode={52242}, 
            state={Iowa},
            country={USA}}

\affiliation[inst2]{organization={Department of Chemical and Biochemical Engineering, Iowa Technology Institute, University of Iowa},
            addressline={4133 Seamans Center}, 
            city={Iowa City},
            postcode={52242}, 
            state={Iowa},
            country={USA}}

\affiliation[inst3]{organization={Pleasant Valley High School},
            addressline={604 Belmont Road}, 
            city={Bettendorf},
            postcode={52722}, 
            state={Iowa},
            country={USA}}

\date{January 2024}

\begin{document}
\doublespacing

\begin{frontmatter}

\begin{abstract}

This study explores the integration of artificial intelligence (AI), specifically Reinforcement Learning (RL), in agricultural management to improve crop yield, optimize nitrogen fertilization and irrigation strategies, and mitigate nitrate leaching and Greenhouse Gas emissions, with a focus on soil Nitrous Oxide (N2O). To address challenges such as climate variability and incomplete knowledge of the agricultural environment, Partially Observable Markov Decision Processes (POMDPs) are adopted to model the interaction between intelligent agents and the agricultural environment using a crop simulator. We employ deep Q-learning, a model-free RL approach, with Recurrent Neural Network (RNN)-based Q networks to train intelligent agents for optimal policies. Additionally, Machine Learning (ML) models are developed to predict N2O emission and seamlessly incorporated into the crop simulator. The research addresses two main areas of uncertainty: the estimation of N2O emissions through a probabilistic ML model and the incorporation of weather condition variability through a stochastic weather generator. Instead of providing a single prediction, the probabilistic ML model offers a nuanced understanding by presenting a range of possible N2O emission outcomes. This method of quantifying uncertainty enhances the model's confidence in its forecasts, enabling more precise decision-making. Moreover, considering climate variability enhances the investigation into the agent's adaptability to climate change, thereby promoting increased resilience of agricultural communities. The results demonstrate the agents' ability to balance crop yield with environmental impacts by penalizing N2O emissions in the reward function. Notably, optimal policies demonstrate adaptability to climate variability, particularly in response to rising temperatures and reduced rainfall. This approach not only optimizes agricultural management against the impact of climate change but also underscores the potential of AI in achieving sustainable farming practices.

\end{abstract}

\begin{keyword}
%% keywords here, in the form: keyword \sep keyword
Agricultural Management \sep Reinforcement Learning \sep Partially Observable Environments \sep Soil N2O Emission \sep Climate Variability \sep Uncertainty
%% PACS codes here, in the form: \PACS code \sep code
%\PACS 0000 \sep 1111
%% MSC codes here, in the form: \MSC code \sep code
%% or \MSC[2008] code \sep code (2000 is the default)
%\MSC 0000 \sep 1111
\end{keyword}
\end{frontmatter}

\section{Introduction}

The escalating challenge of climate change, profoundly impacting global ecosystems, requires immediate and innovative solutions. Greenhouse Gases (GHGs) play a crucial role in climate change by trapping heat in the atmosphere. Nitrous Oxide (N2O), a primary GHG, is produced by both natural and human-induced processes, particularly through nitrogen-based fertilizer use and other farming practices. Simultaneously, climate variability poses a formidable threat to agricultural productivity, jeopardizing food security worldwide. According to the Food and Agriculture Organization (FAO) data, approximately 828 million people still experienced hunger in 2022. Agriculture, a vital component of the global economy, faces a dual challenge - navigating the impacts of GHGs on climate change and addressing the threats posed by climate variability. This intricate interplay underscores the need for a paradigm shift in agricultural management.

In past research on agricultural management, scholars typically gathered and examined historical data to uncover crop growth patterns. These findings were then used to guide future agricultural policies and practices~\cite{Kenneth2002}. However, with the continuous advancement of computer hardware and simulation software, there has been a notable shift in research methodologies. Specialized software tools, such as Decision Support System for Agrotechnology Transfer (DSSAT) \cite{Jones2003}, Agricultural Production Systems Simulator (APSIM) \cite{Keating2003}, and AquaCrop \cite{Steduto2009}, have been developed and widely adopted in the agricultural research community. These simulation tools encompass various aspects of crop development, yield, water, and nutrient needs, enabling the optimization of management practices to adapt to evolving weather and environmental conditions.  

With the rising interest in Artificial Intelligence (AI) for smart or precision agriculture \cite{Zhang2022}, researchers are increasingly integrating AI techniques, including Reinforcement Learning (RL), with the established software mentioned above to simulate and formulate improved agricultural management strategies. As a subset of Machine Learning (ML), RL empowers computer programs, functioning as agents, to navigate unfamiliar and dynamic systems for specific tasks \cite{Li2023, Cai2023}. Romain \textit{et al.} \cite{Romain2022} transformed DSSAT into a realistic simulation environment suitable for RL, known as Gym-DSSAT, which has gained popularity in agricultural research. Wu \textit{et al.} \cite{Wu2022} demonstrated that RL-trained policies could outperform traditional empirical methods, achieving higher or similar crop yields while using fewer fertilizers, a significant advancement in sustainable agricultural practices. Complementing this, Sun \textit{et al.} \cite{Sun2017} explored RL-driven irrigation control, optimizing water usage while maintaining crop health and showcasing the potential of Gym-DSSAT in effective resource management. Furthermore, Wang \textit{et al.} \cite{Wang2024} verified the robustness of learning-based fertilization management under challenging conditions. Even in extreme weather scenarios, the RL agent demonstrated the ability to learn optimal policies, resulting in highly satisfactory outcomes. This underscores the reliability and adaptability of RL in varying environmental conditions.

Most existing studies \cite{Romain2022, Wu2022, Sun2017} have predominantly assumed a completely observable agricultural environment, formulating the related RL problems as Markov Decision Processes (MDPs). In MDP frameworks, it is assumed that each state of the environment contains all the necessary information for the agent to identify the optimal action for achieving the objective function. However, a significant issue arises when mirroring real-world scenarios, where agents lack complete knowledge to accurately determine the state of the environment due to the often uncertain or partial nature of their observations\cite{Williams2022}. Notably, certain state variables in Gym-DSSAT, such as the index of plant water stress, daily nitrogen denitrification, and daily nitrogen plant population uptake, may pose challenges in terms of measurements and accessibility. Wang \textit{et al.} \cite{Wang2024} delved into this issue and discovered that it can be effectively addressed through the application of Partially Observable Markov Decision Processes (POMDPs). Subsequently, they adopted Recurrent Neural Networks (RNNs) to handle the history of observations for decision-making in fertilization management. Their findings indicated that modeling the agricultural environment as a POMDP resulted in superior policies compared to the existing assumption of an MDP.

Across the globe, the upward trend in N2O emissions, both historically and in projections, is primarily attributed to the expanding use of fertilizers and the growth in livestock production. Approximately 60\% of the contemporary increase in N2O comes from cultivated soils receiving Nitrogen (N) fertilizers \cite{IPCC}. Notably, from 1990 to 2020, there has been a 34.9\% increase in N2O emissions from agricultural soils \cite{EPAN2O}. Various factors can influence N2O emissions, including crop types, tillage methods, crop residue management strategies, soil moisture levels, soil temperature conditions, and aspects of fertilizer usage. These aspects encompass the quantity, type, application timing, and method of placement \cite{Shcherbak2014}. In addition to anthropogenic factors, climate variability also plays a pivotal role in agricultural management, considering fluctuations in temperature, rainfall, wind patterns, and other weather elements across different time and space scales \cite{IPCC2014}. 

This research harnesses advanced AI techniques, specifically ML and RL, to foster optimal agricultural management for sustainable crop production while mitigating environmental impact, notably soil N2O emissions. The utilization of Gym-DSSAT, a robust crop simulator, is complemented by the adoption of POMDP to model the interaction between agents and the agricultural environment. Our approach involves developing deterministic and probabilistic ML models seamlessly integrated with the crop simulator. Considering the challenges of partially observable agricultural environments, we employ RNN-based deep Q-learning, a model-free RL method, to train the agents for optimal policies. The research extends its focus to investigate the adaptability of optimal policies in response to climate variability, particularly emphasizing temperature rise and rainfall reduction. To address weather uncertainties, we incorporate a stochastic weather generator. This comprehensive methodology positions our study at the forefront of advancing sustainable agricultural practices by integrating cutting-edge AI technologies. 

In this study, we present the inaugural effort to make a significant contribution to bridge the gap in understanding the mutual effects between agricultural management strategies, specifically fertilization and irrigation plans, and challenges posed by climate change. By incorporating predicted soil N2O emissions into the reward function, the developed RL method can successfully guide agents in learning farming practices to mitigate GHG emissions, with a particular focus on N2O emissions. This approach, coupled with other considerations, provides valuable insights into fostering more sustainable agricultural practices. 

\begin{comment}
In this study, we have one contribution toward managing the environmental impacts of agriculture, particularly focusing on soil N2O emissions. Our work bridges the gap between agricultural management strategies, specifically in the areas of fertilization and irrigation, and their consequent effects on soil N2O emissions. By using RL, we have been able to delineate how various fertilization and irrigation practices contribute to the release of N2O, a potent GHG, thus providing insights into more sustainable agricultural practices.

Our study also makes a significant contribution by exploring and measuring the effects of various fertilization and irrigation methods on crop yield and soil nitrous oxide (N2O) emissions in the context of unpredictable weather conditions. Through our simulations, we have gained insights into how weather variability impacts the efficiency of different fertilization and watering approaches, as well as their outcomes on yield and N2O emissions. As far as we are aware, there have been no reported systematic studies in the literature concerning learning-based approaches to agricultural management. 
\end{comment}

Another significant contribution is the enhanced uncertainty quantification of optima policies' performance, representing a progression from our prior study \cite{Wang2024}. By integrating a probability ML model for N2O emission prediction and a stochastic weather generator into the crop simulator, our agents exhibit the capability to learn adaptive optimal policies for fertilization and irrigation, particularly responding to climate variability, including rising temperature and reducing rainfall. This adaptation extends to address severe climate events like droughts. 
 
This paper is organized as follows: Section 2 introduces the formulations of POMDP and discusses Deep Q-learning, a model-free RL technique. In Section 3, we establish the ML models for N2O emissions and outline the simulation model settings. Section 4 explores the integration of N2O emissions into the management of N fertilization and irrigation while also examining the implications of weather variability, such as elevated temperatures and reduced precipitation. The paper concludes with Section 5, where we briefly summarize our findings, engage in discussion, and propose alternative solutions for future research.

\section{Methodologies}

In this study, we conceptualize the agricultural environment as a POMDP and employ a model-free RL method for the agent to acquire optimal policies. This section begins by establishing the mathematical framework of POMDP. Following that, we introduce Q-learning and its variations, emphasizing their relevance in addressing POMDP-related challenges.   

\subsection{POMDP}

A POMDP is usually represented by a tuple $\mathcal{P}=\left(S, A, T, s_0, R, O, \Omega \right)$, including a finite set of states $S = \{s_1, ..., s_n\}$, a finite set of actions $A=\{a_1, ..., a_m\}$, the initial state $s_0 \in S$, and a finite set of observations $O = \{o_1, ..., o_q\}$. Particularly, $A(s)$ is a set of available actions at state $s$ for the agent to take. When the agent takes an action $a \in A(s)$, a transition occurs from the current state $s$ to the next state $s'$ with a probability $T(s,a,s')$. Such transition probability is denoted by a function $T : S\times A\times S\rightarrow \left[0,1\right]$ and satisfy $\sum_{s'\in S} T(s,a,s') = 1$. 

After each transition, the agent may receive feedback based on the reward function $R: S \times A \times S \rightarrow R$. In addition to $R(s, a,s')$, the reward function has various formulations like $R(s')$ and $R(s,a)$. Since the environment is partially observable, a set of possible observations the agent can perceive is defined as $O(s')$. There exists an observation probability function $\Omega: S \times A \times O \rightarrow \left[0,1\right]$ to quantify the perception uncertainty after the agent takes action $a$ and reaches the next state $s'$. This function must satisfy $\sum_{o\in O} \Omega(s',a,o) = 1$.

The primary goal of the agent in an RL problem is to learn an optimal policy that can maximize the expected return, also known as the utility. Beginning from the current state $s$ and adhering to a policy $\xi$, the expected return is the accumulated rewards the agent can collect. It is defined below as the sum of discounted rewards over a sequence of interactions with the environment.

\begin{equation}\label{eq:expReturn} 
U^{\xi}(s) = \mathbb{E}^{\xi} \left[\sum_{t=0}^{\infty} \gamma^t R(s_t, a_t, s_{t+1}) \Big \vert s_{t=0} = s\right]
\end{equation}
where $s_{t}$ represents the state of the environment at time $t$, and $a_t$ is the action to be taken, potentially leading to the transition of the agent to the state $s_{t+1}$ at the next time, $t+1$. The discount factor, $\gamma \in [0,1]$, is commonly employed to weigh the importance of future rewards in the agent's decision-making process. The utility in Equation (\ref{eq:expReturn}) assesses the expected total reward an agent can accrue in the long run and is also referred to as the state value, denoted as $V(s)$.

 \subsection{Q-Learning}

Q-learning \cite{Watkins1992} is a widely used model-free RL method, and it utilizes Q values (action values or state-action values) to evaluate and select actions during the learning process. Similar to state values, the Q value, denoted as $Q^{\xi}(s,a)$, represents the total reward an agent is expected to accumulate after taking action $a$ at the state $s$ while following a policy $\xi$. Q values and state values are related through $V(s) = {\mathrm{max}}_{a} Q(s,a)$. In contrast to policy-based RL methods \cite{Zhu2022}, value-based methods like Q-learning directly seek the optimal value functions. These functions are subsequently used to select actions through the greedy technique. The $\varepsilon$-greedy is usually adopted during the learning process to balance exploration and exploitation. 

Given that the agricultural management problems under study involve an infinite state space, traditional tabular Q-learning is not suitable. Therefore, we adopt deep Q-learning, also known as deep Q networks or DQN \cite{Mnih2013}, where Q values are approximated by Deep Neural Networks (DNNs). DQN employs two DNNs with identical network architectures. However, only one DNN, referred to as the evaluation Q-network, is trained and updated with collected experiences at every step. The other DNN, known as the target Q-network, periodically copies the weight of the evaluation Q-network.   

\begin{figure}
\centering
\resizebox*{10cm}{!}{\includegraphics{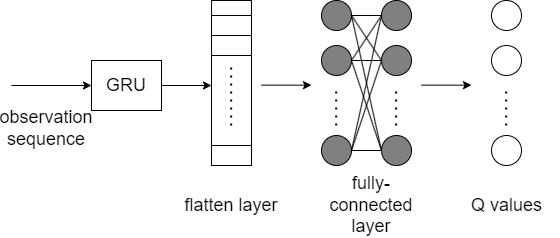}}
\caption{GRU-based Q-network architecture.} \label{fig:QNet}
\end{figure}

Moreover, our prior work \cite{Wang2024} demonstrated that the agricultural environment is better modeled as POMDP rather than MDP. This is because the agent cannot completely identify the state of the agricultural environment based on limited observations. In POMDPs, decision-making relies on the history of observations instead of the current one. To address this challenge, we incorporated a Recurrent Neural Network (RNN) like Gated Recurrent Unit (GRU) \cite{Cho2014} into the Q network architecture, as shown in Figure \ref{fig:QNet}. This allows the Q-networks to take the observation sequence as input and approximate Q values as $Q(\mathbf{o}_t, a_t)$ where $\mathbf{o}_t$ represents the history of observations up to time $t$.

As a result, the two Q-networks in our DQN are denoted as $Q_E(\mathbf{o}_t, a_t;\theta_E)$ and $Q_T(\mathbf{o}_t,a_t; \theta_T)$ where $\theta_E$ and $\theta_T$ are network weights for the evaluation and target Q-networks, respectively. In each step of the learning process, the agent selects an action $a_t$ at the current state $s_t$ based on the Q values predicted from the evaluation Q-network, with the observation history $\mathbf{o}_t$ as the input. The $\varepsilon$-greedy technique is employed for action selection. 

Following the execution of the action and the transition to the next state $s_{t+1}$, the agent receives a reward $r_t = R(s_t, a_t, s_{t+1})$, observes an observation $o_{t+1}$, and generates a new observation sequence $\mathbf{o}_{t+1} = (o_{t-l+2}, o_{t-l+3}, ..., o_{t+1})$ with a length of $l$. Simultaneously, the experience, represented as $(\mathbf{o}_t, a_t, r_t, \mathbf{o}_{t+1}))$, is stored in the experience replay memory \cite{Lin1992}. Each experience contributes to one data sample, updating the Q value associated with the observation sequence and the action taken through the Bellman equation \cite{Sutton2018}.
\begin{equation}\label{eq:DRQN Q value}
Q_{new}(\mathbf{o}_t, a_t) = Q_E(\mathbf{o}_t,a_t; \theta_E)+\alpha \left[ r_t+\gamma \max_{a_{t+1}} Q_T(\mathbf{o}_{t+1},a_{t+1}; \theta_T)-Q_E(\mathbf{o}_t,a_t; \theta_E) \right]
\end{equation}
where $\alpha$ is the learning rate. 

At each time step, a batch of data samples is randomly selected to train and update the evaluation Q-network. Meanwhile, the target Q-network maintains constant weights until it copies from the evaluation Q-network, i.e., $\theta_T = \theta_E$, after a certain number of time steps. 

\section{Model setup}

The simulation model generated in this study aligns with the Long-Term Ecological Research (LTER) site at the W.K. Kellogg Biological Station (KBS-LTER; 42${^\circ}$ 24' N, 85${^\circ}$ 24' W, 288 m elevation), as established in 1989 \cite{robertson and hamilton 2015}. The testing field follows a no-till corn-soybean-winter wheat rotation and contains 1.6\% solid organic carbon. The climate at this site is classified as humid continental, characterized by a mean annual precipitation of 1151 mm and an average temperature of 7.6 $^\circ$C. For more information on agronomic management details, please refer to the KBS-LTER data tables available in \cite{KBSData}. 

\subsection{N2O emission forecasting}

Given that the Gym-DSSAT platform lacks N2O emission forecasting capabilities, this study endeavors to fill the gap by developing ML models, deterministic or probabilistic. These models aim to predict N2O emissions based on a combination of weather conditions and agricultural management practices. The dataset we used is from Saha's study \cite{Saha2021}, spanning the years 2012 to 2017 (excluding 2015 due to instrument failure). The dataset contains numerous features. However, to align with the state variables available in Gym-DSSAT, we select four specific features, as outlined in Table ~\ref{table:N2Oinput}. The model's output is expressed in grams of nitrogen emitted per hectare each day (g N2O-N/ha/d). The dataset comprises a total of 919 samples. For training and testing purposes, 80\% of these samples are allocated to training and 20\% to testing. To ensure robust validation, we employ a 5-fold cross-validation method. 
 
\begin{table}
\centering
%\begin{tabular} {||c c||} 
\begin{tabular} {| p{2.5cm} | p{10cm} | }
 \hline
 Variable & Description \\ [0.5ex] 
 \hline\hline

 \hline
 \textbf{pp2} & the total precipitation in the 2 days leading up to gas sampling (mm)\\
 \hline
 \textbf{pp7} & the total precipitation in the 7 days leading up to gas sampling (mm)  \\
 \hline
 \textbf{airT} & the mean daily air temperature ($^\circ$C)  \\ 
 \hline
 \textbf{daysAF} & the number of days that have elapsed following the application of top-dressed nitrogen fertilizer \\ 
 \hline
 \end{tabular}

\caption{Features for N2O emission forecasting models. }\label{table:N2Oinput}
\end{table}

\begin{figure}
\centering
\resizebox*{9cm}{!}{\includegraphics{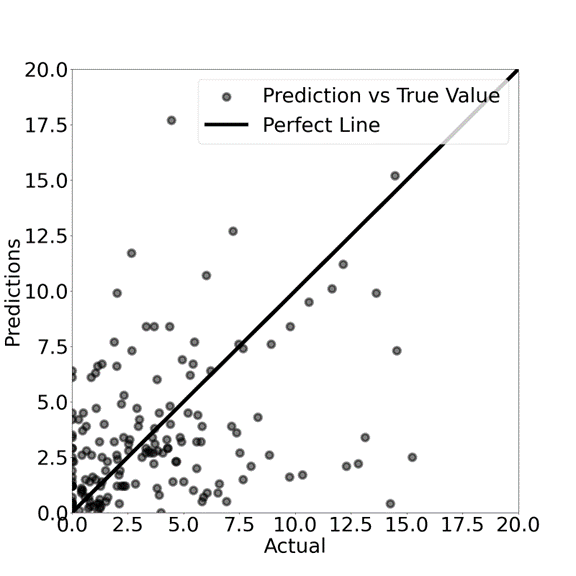}}
\caption{The predictive N2O daily flux (g/ha) compared to true values by using a deterministic ML model. } 
\label{fig:N2OANN}
\end{figure}

The first ML model employed for N2O prediction is a deterministic ML model with an artificial neural network (ANN). The neural network architecture comprises four layers, each consisting of 512 neurons with Rectified Linear Unit (ReLU) activation functions. Training involves a batch size of 128, a learning rate set at 0.0001, and a total of 6,000 epochs. The performance of the model on the testing set is visually presented in Figure~\ref{fig:N2OANN}, showcasing a comparison between predictive and true values.

Despite an extensive training regimen, the final coefficient of determination ($R^2$) achieved for the testing set stands at 0.65, indicating a moderate level of predictive accuracy. It is noteworthy that we conducted experiments with various neural network architectures and activation functions, and the configuration described above yielded the best performance. While the $R^2$ score may not reach anticipated highs, it's worth emphasizing that our model's performance closely aligns with the outcomes observed in Saha's study, which reported an approximate $R^2$ of 0.67 \cite{Saha2021}.

The deterministic ML model described earlier provides a singular optimal prediction for N2O emissions. Utilizing the sum of squared residuals as the loss function, a common practice in least square regressions, this model aims to minimize the difference between predicted and observed values. However, recognizing the inherent data uncertainty arising from measurements in the testing field, we take a different approach for training a probabilistic ML model - utilizing the maximum likelihood (MaxLike) method \cite{Myung2003}. Unlike its deterministic counterpart, this model doesn't offer a single-point prediction but rather predicts a probability distribution, encompassing all potential N2O emissions. 

MaxLike estimation is commonly employed to identify a suitable probability distribution with parameters that best explain data samples. Consequently, training a probabilistic ML model becomes a probability density estimation problem. This involves searching for optimal model parameters, denoted as $\theta$, with the objective of maximizing the joint probability of a given dataset $(X, y)$ where $X=(x_1, ..., x_n)$ and $y=(y_1, ..., y_n)$. The joint probability is often expressed as a likelihood function, denoted as 
\begin{equation}
L(y|X;\theta)=P(y_1, ..., y_n | x_1, ..., x_n; \theta) =  \prod_{i=1}^n P(y_i | x_i; \theta)
\end{equation}
where the data samples are assumed to be independent and identically distributed, so the likelihood function can be reformulated as the multiplication of conditional probabilities.

As multiplying numerous small probabilities together can be numerically unstable in practice, using the sum of log conditional probabilities is common. Consequently, the Negative Log-Likelihood (NLL) function is typically employed as the cost function, as shown below. This function is minimized during the training of a probabilistic ML model.
\begin{equation}
\textrm{min(NLL)} = \textrm{min} -\sum_{i=1}^n logP(y_i | x_i; \theta)
\end{equation}

On the other hand, given that the N2O emission cannot be negative, we chose the log-normal distribution over the normal distribution, expressed as $ln(X) \sim \mathcal{N}(\mu, \sigma^2)$. The probability density function is defined as 
\begin{equation}\label{eqn:Lnormal}
P(y_i| x_i; \mu_{x_i}, \sigma_{x_i}) = \frac{1}{y_i \sigma_{x_i} \sqrt{\pi}} \textrm{exp}\left(-\frac{(ln(y_i) - \mu_{x_i})^2}{2 \sigma^2_{x_i}}\right)
\end{equation}
where $\mu_{x_i}$ is the location parameter, and $\sigma_{x_i}$ is the scale parameter. 

\begin{figure}
\centering
\resizebox*{9cm}{!}{\includegraphics{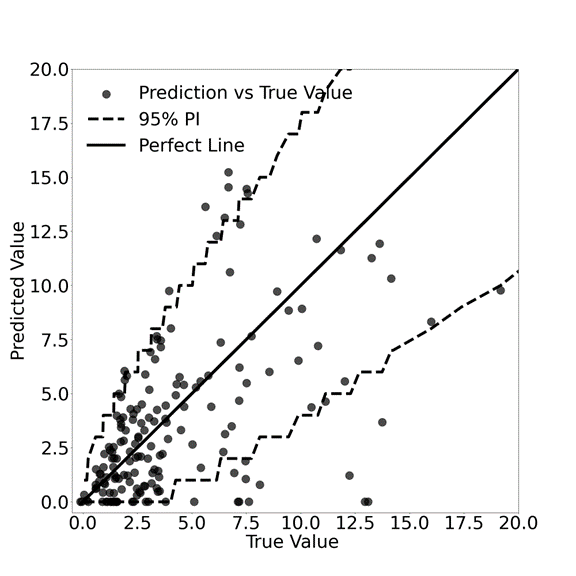}}
\caption{The predictive N2O daily flux (g/ha) compared to true values by using a probabilistic ML model.} \label{fig:N2OPDL}
\end{figure}

We employ another ANN for the probability ML model, featuring a four-hidden-layer architecture with 16, 32, 64, and 16 neurons, respectively. Diverging the deterministic model, the output layer of this model consists of two neurons: one for $\mu_x$ and the other for $\sigma_x$, representing the parameters of the Log-normal distribution in Eqn.~(\ref{eqn:Lnormal}). The training process spans 5000 epochs, with a batch size of 16. For configuration and training, we leverage the Tensorflow-probability package \cite{Durr2020}. Figure \ref{fig:N2OPDL} visually illustrates the model's performance using the testing data. Each prediction, sampled from the predictive probability distribution, is compared to the corresponding true value or observation. The figure also includes a 95\% prediction interval (PI), providing a comprehensive assessment of the model's performance by considering not only its central tendency but also its variability. 

It is important to note that the collected data \cite{Saha2021} includes an extra input feature: the amount of N fertilization. However, the dataset only provides a single recorded value for this input feature, specifically 170 kg/ha. To estimate N2O emission across varying N input amounts, we employ regression analysis derived from Hoben's exponential model \cite{Hoben2011}. The resulting approximation is expressed as $y(x) = y(170) \cdot e^{0.0073 \cdot (x - 170)}$ where $x$ represents the actual N input, and $y(170)$ denotes the average daily N2O flux predicted from the ML models under the assumption of an N input of 170 kg/ha. Additionally, the sum of rainfall and irrigation is considered to calculate the precipitation-related input features (refer to Table~\ref{table:N2Oinput}) for predicting N2O emission. 

\subsection{Crop simulator}

This study employs Gym-DSSAT as a crop simulator, facilitating the approximation of interactions between the agent and the agricultural environment. Through RL methods, the agent learns optimal agricultural management, also called optimal policies. The Gym-DSSAT encompasses 28 internal variables. Previous studies \cite{Wu2022, Sun2017} commonly used all these variables as state variables, assuming completely observable agriculture environments. Consequently, the learned policies in those studies mapped current observations directly to the management plan. In other words, the agent made decisions regarding fertilization and irrigation based on current observations of state variables.

In contrast to conventional approaches, we recognize the practical challenges associated with measuring certain variables. As highlighted in our previous study \cite{Wang2024}, we consider the agricultural environment to be partially observable. In our model, intelligent agents base decisions on the historical observations of a carefully selected set of ten state variables, outlined in Table~\ref{table:stateVar}. This nuanced perspective aims to offer a more realistic representation of the agent's decision-making environment, ensuring alignment with the complexities found in real-world scenarios. 

\begin{table}
\centering
%\begin{tabular} {||c c||} 
\begin{tabular} {| p{2.5cm} | p{10cm} | }
 \hline
 Variable & Description \\ [0.5ex] 
 \hline\hline

 \hline
 \textbf{cumsumfert} & cumulative nitrogen fertilizer applications (kg/ha) 
 \\ \hline
 \textbf{dap} & days after planting  \\
 \hline
 \textbf{istage} & DSSAT maize growing stage  \\ 
 \hline
  \textbf{pltpop} & plant population density (plant/m$^2$) \\ 
 \hline
 \textbf{rain} & rainfall for the current day (mm/d)	 \\
 \hline
 \textbf{sw} & volumetric soil water content in soil layers (cm$^3$ [water] / cm$^3$ [soil]) \\
 \hline
 \textbf{tmax} & maximum temperature for the current day ($^{\circ}$C)  \\
 \hline
 \textbf{tmin} & minimum temperature for the current day ($^{\circ}$C) \\
 \hline
 \textbf{vstage} & vegetative growth stage (number of leaves)   \\
 \hline
 \textbf{xlai} & plant population leaf area index   \\
 \hline
 \end{tabular}

\caption{State variables of the agricultural environment used in this study. }\label{table:stateVar}
\end{table}

In this study, the action space encompasses different combinations of N and water quantities that can be applied in a single day. Mathematically, the action space is discretized as $(N_p, I_q)$ where $N_p = 20p (kg/ha)$ represents the N input, and $I_q = 10q (L/m^2)$ represents the water input. Both $p$ and $q$ vary within the range of 0 to 4. Consequently, there are a total of 25 available actions for the agent to choose from each day.

On a given day $d_t$, after the execution of a selected action involving the application of N input $N_t$ and water input $I_t$, Gym-DSSAT conducts computation for nitrate leaching $L_t (kg/ha)$ and crop yield $Y (kg/ha)$ if harvested. Additionally, our ML models estimate N2O emissions $O_t (kg/ha)$. Following these calculations, the agent is rewarded according to the formula specified in Eqn~(\ref{eq:AgReward}).

\begin{equation} \label{eq:AgReward}
R_t= \left\{ \begin{array}{cc} w_1 Y - w_2 N_t - w_3 I_t - w_4 L_t - w_5 O_t & \mbox{at harvest}
\\ -w_2 N_t - w_3 I_t - w_4 L_t - w_5 O_t & \mbox{otherwise} \end{array} \right.
\end{equation}
where $w_1=0.2$, $w_2=2$, $w_3=2$, $w_4=30$, and $w_5=100$ represent the weight coefficients. It is important to highlight that $w1$ through $w3$ align with those utilized in previous research studies \cite{Tao2022}. We explored alternative values for the weight assigned to nitrate leaching and N2O emissions in the reward function and assessed the resulting outcomes. The comparative analysis for the year 2012 indicated that a weight combination of 30 and 100 yields superior results, ensuring optimal output levels while maximizing the reward. 

Our research focuses on the growth and yield of corn for the year 2012, requiring the extraction of climate and soil conditions specific to that timeframe. The relevant meteorological data was obtained from the KBS-LTER website \cite{KBSData}, offering detailed daily records of maximum and minimum temperatures, precipitation, and solar radiation. To address the variability in climatic conditions, we utilized the stochastic Weather Generator (WGEN) \cite{WGEN2003}, a random weather generator integrated into DSSAT. This tool enables the generation of weather scenarios for each episode under investigation. 

The WGEN categorizes its output variables into two distinct groups. The first group exclusively encompasses precipitation, while the second group comprises maximum temperature, minimum temperature, and solar radiation. This categorization is based on the understanding that the occurrence of rain on a given day significantly influences that day's temperature and solar radiation. As a result, precipitation is generated as an independent variable each day, separate from the other variables in the second group. Then, calculations for maximum and minimum temperatures and solar radiation are executed depending on whether the day is characterized as wet or dry. 

More specifically, the WGEN incorporates a precipitation element based on a Markov chain-gamma distribution model. It employs a first-order Markov chain model to predict the likelihood of rain, considering whether the previous day was wet or dry. In the case of a predicted wet day, a two-parameter gamma distribution is utilized to calculate the precipitation amount. Subsequently, the residuals for the other three variables - maximum temperature, minimum temperature, and solar radiation - are generated through a multivariate normal generation process. This process maintains the serial and cross-correlation coefficients of the variables. The final values for these three variables are determined by adding the calculated residuals to the seasonal means and standard deviations, following the methodology outlined in \cite{Richardson1985}. 

\section{Simulation results and discussions }

In this study, we employ an RNN-based DQN, as detailed in Section 2.2, to facilitate the agent in learning optimal policies. The Q-networks, integrated into this approach, take a sequence of observations as input and generate Q values, guiding the agent in its action selection. The RNN layer within the Q-networks (refer to Figure~\ref{fig:QNet}) consists of a single hidden layer with 64 units. Its output is subsequently fed into a fully connected network. Our study utilizes a sequence comprising observations from five consecutive days to make decisions. 

Throughout the learning process, we apply the $\varepsilon$-greedy selection technique to strike a balance between exploration and exploitation. The discount factor, crucial for future reward, is set at 0.99. To design and update the neural networks, we employ PyTorch and Adam optimizer \cite{Kingma2014}, using an initial learning rate of 1e-5 and a batch size of 640. The choice of parameters is based on considerations of model performance and efficiency. Simulations are conducted on two distinct machines. The first machine is equipped with an Intel Core i7-12700K processor, an NVIDIA GeForce RTX 3070 Ti graphics card, and 64GB RAM. The second machine features an AMD 5800h processor, an NVIDIA GeForce RTX 3070 graphics card, and 32GB of RAM. The selection of these configurations is informed by their computational capabilities and relevance to the scope of our study. 

We conduct multiple simulations to explore the implications of N2O emission and climate variability on agricultural management and outcomes, particularly corn yield. First, we select the year 2012 as our baseline, utilizing authentic weather data and soil properties. By incorporating N2O emission into our reward function, we simulate the effects of N2O emission in the context of agricultural practices. Following this, we introduce variations in temperature and rainfall to assess the influence of climate variability. To enhance the model's resilience against unpredictable weather conditions, we choose to generate randomized weather scenarios based on actual data using the WGEN, as described in Section 3.2. These scenarios are then utilized to train our models, significantly enhancing their accuracy in coping with uncertain environmental conditions.

\subsection{Considering N2O emission }

In this task, our objective is to investigate the influence of N2O emission on learned management policies and agricultural outcomes through simulations in three distinct cases. The first case focuses solely on nitrate leaching, excluding consideration of soil N2O emissions. In the second, we shift the focus to concentrate exclusively on soil N2O emissions, neglecting the impacts of nitrate leaching. Finally, we integrate both factors in the third case, analyzing the simultaneous effects of nitrate leaching and N2O emissions. It is worth noting that we utilize the deterministic ML model to predict daily N2O emissions in the above-mentioned cases. 

Upon comparing the actual data received from KBS \cite{KBSData}, we gather all available information, encompassing diverse fertilization and irrigation practices across various testing fields from 2011 to 2014. The N inputs exhibit significant variation, ranging from 0 kg/ha to a maximum of 291 kg/ha, with an average input of 138 kg/ha. Yield outcomes also display variability, with the highest recorded yield at 14023 kg/ha, the lowest at 3084 kg/ha, and an average yield of 9740 kg/ha. Additionally, the measured N2O daily fluxes reach up to 0.6 (kg/ha/d) with a median value of 0.002. Unfortunately, detailed irrigation data are not available. 

In our simulations, following the acquisition of the optimal policy via RL in each case, we apply it to perform one realization in the year 2012. The comparative results across the three cases are presented in Table~\ref{table:DiffwtCompare}. The total rewards exhibit similarities in the three cases, as do the yields, falling within the range of actual data and surpassing the average (9740 kg/ha). 

Data from The Mosaic Company \cite{Mosaic} indicates that a corn crop yielding 200 bushels per acre (equivalent to 12,553 kilograms/hectare) can absorb up to 297kg of N per hectare. When N inputs align with the crop's requirements, there is no noticeable increase in N2O emissions. As N inputs exceed the crop's needs, N2O emissions begin to rise dramatically \cite{Shcherbak2014}. It is worth noting that the N inputs in all three cases fall below the mentioned threshold. Furthermore, by considering N2O emission in the reward function, the resultant policies have the potential to mitigate N2O emissions in Case 2 and reduce both nitrate leaching and N2O emission in Case 3, all while maintaining production levels.

\begin{table}
\centering 
\begin{tabular} {| p{4.5cm} | p{2cm} | p{2cm} | p{3cm} |}

 \hline
 & Case 1 & Case 2 & Case 3 \\ [0.5ex] 
 \hline\hline

 \hline
 Reward  & 1338  & 1267  & 1272
 \\ 
 \hline
 Yield($kg/ha$)  &  11,190 & 10,549  & 11,305 
 \\
 \hline
 N input($kg/ha$)  & 140  & 140  & 180 
 \\
 \hline 
 Water input($L/m^2$) & 310  & 270 & 300
 \\
 \hline
 Nitrate leaching($kg/ha$) & 0.0009 & 0.21 & 0.175
 \\
 \hline
 N2O emission($kg/ha$) & 0.314 & 0.223 & 0.251
 \\
 \hline

\end{tabular}

\caption{Agriculatural outcomes for three difference cases (Case 1: Nitrate leaching only, Case 2: N2O emission only, and Case 3: Nitrate leaching and N2O emission.)}\label{table:DiffwtCompare}
\end{table}

The total fertilizer and water usage quantities are relatively comparable, but the application policies differ. This disparity arises because the agent strategically balances gains, like corn yield, against costs, including fertilizer and water usage, as well as penalties for nitrate leaching, N2O emission, or both, all with the goal of maximizing the overall reward.

\begin{figure}
\centering
\resizebox*{11cm}{!}{\includegraphics{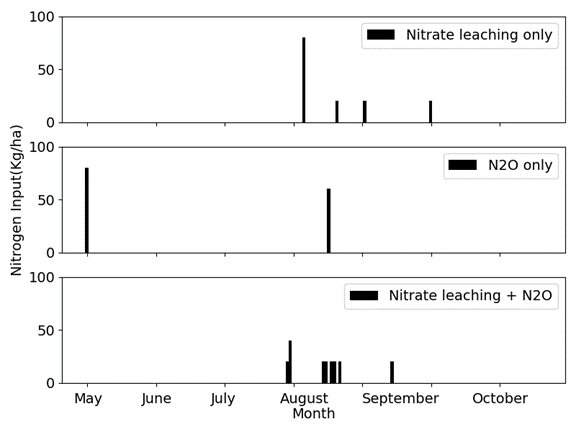}}
\caption{Fertilization strategies across three cases.} \label{fig:N2OF}
\end{figure}

\begin{figure}
\centering
\resizebox*{11cm}{!}{\includegraphics{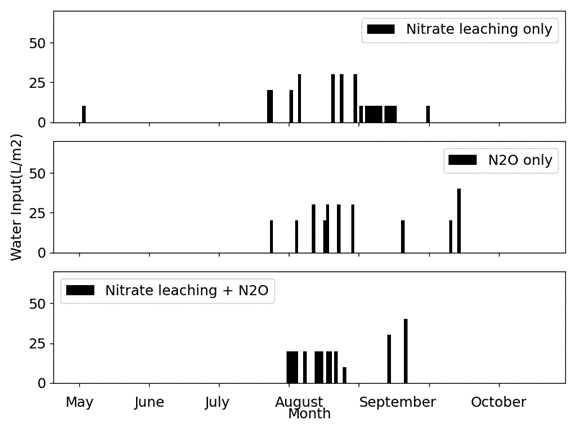}}
\caption{Irrigation strategies across three cases.} \label{fig:N2OI}
\end{figure}

Figures~\ref{fig:N2OF} and~\ref{fig:N2OI} provide detailed insights into various fertilization and irrigation strategies (i.e., plans) based on optimal policies learned in three distinct cases. Notably, fertilizers and water are primarily applied in August and September, crucial growth months for corn. Furthermore, an analysis of the 2012 weather data reveals a significant reduction in precipitation during these two months, approximately 25\% less than in preceding years. This reduction coincided with a rise in average temperatures compared to historical records \cite{weather}, emphasizing the urgent need for irrigation.

Particularly in Case 2, we observe a decrease in the frequency of fertilization, with one fertilization event occurring independently from irrigation applications. The total water usage in this case is also slightly lower than in the other two cases. These factors collectively contribute to a reduction in N2O emissions, aligning with the findings of Weitz \textit{et al.} \cite{Weitz2001}. Their research emphasizes that soil moisture dynamics significantly influence N2O emissions, noting the highest post-fertilization emissions observed in moist soil. Emissions from drier soils only increase following rainfall. 

\subsection{Temperature rising}

Significant shifts in global climate patterns have been marked by a rise in air temperature \cite{Rabatel2013}, primarily attributed to GHG emissions resulting from human activities. According to historical data from NASA, there has been a consistent increase in average temperatures since 1880. This trend of global warming has become more pronounced in recent years, with temperatures rising by 0.94 degrees Celsius in the past 60 years \cite{NASA}. 

In this study, we use the year 2012 as the baseline and augment monthly maximum and minimum temperatures by up to 3 degrees Celsius using WGEN to generate random weather. In contrast to our previous investigation \cite{Wang2024}, where the temperature pattern was preserved, the randomly generated weather in this study does not replicate the identical patterns observed in 2012, introducing weather uncertainty through temperature variations. Notably, while rainfall is also randomly generated via WGEN correspondingly, the monthly total precipitation remains the same as observed in 2012. Furthermore, we integrate the probabilistic ML model developed in Section 3.1 to assess N2O emissions, considering data uncertainty from measurements in the testing fields.  

We examine temperature increases of 0.5, 1, 1.5, 2, 2.5, and 3 degrees Celsius. Additionally, we include a scenario with no temperature increase but random weather. Two types of policies are implemented: the "fixed policy," derived from actual weather data of 2012, which considers both nitrate leaching and N2O emission as discussed in Case 3 in the previous subsection, and the "optimal policies," specifically learned at each temperature increase. Following policy learning, 300 realizations are conducted to assess the uncertainties associated with agriculture outputs and management. This approach also allows for an exploration of policy adaptability to climate variability.  

To enhance the training efficiency of the agent in learning optimal policies, we leverage a transfer learning technique - fine-tuning. The evaluation Q network associated with the fixed policy serves as a pre-trained model, acting as the starting point for training optimal policies or updating the evaluation Q network at each specific temperature increase. The adoption of fine-tuning yields a significant reduction in training time compared to conventional methods that start with a random policy. 

\begin{figure}
\centering
\resizebox*{10cm}{!}{\includegraphics{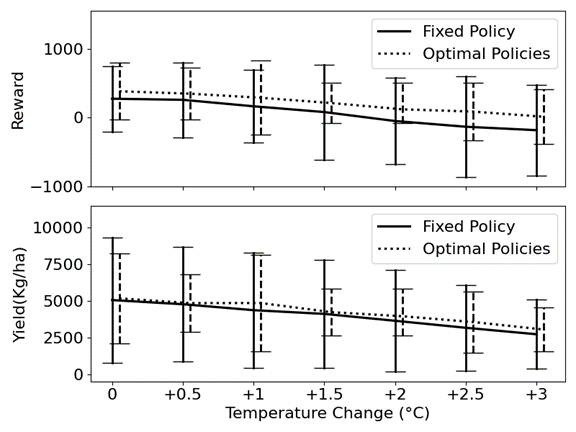}}
\caption{Rewards and corn yields with 95\% PI from different policies when monthly temperature increases. The figure also includes the range of actual corn yield data.} \label{fig:TRY}
\end{figure}

\begin{figure}
\centering
\resizebox*{10cm}{!}{\includegraphics{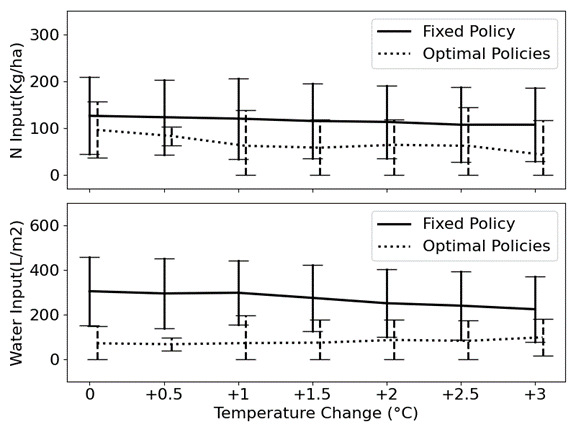}}
\caption{N and water inputs with 95\% PI from different policies when monthly temperature increases.} \label{fig:TNW}
\end{figure}

\begin{figure}
\centering
\resizebox*{10cm}{!}{\includegraphics{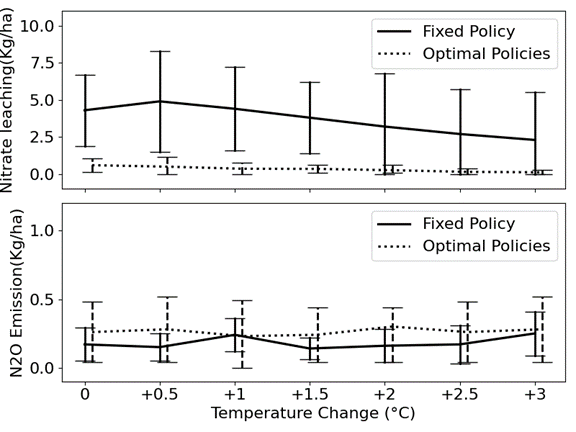}}
\caption{Nitrate leaching and N2O emission with 95\% PI from different policies when Monthly temperature increases.} \label{fig:TNL}
\end{figure}

Figure~\ref{fig:TRY} depicts agricultural outcomes, including corn yields and total rewards, along with 95\% PIs for the fixed and optimal policies. The data shows a consistent decline in average rewards and yields with increasing temperature, underscoring the adverse effects of rising temperatures on agricultural production. Nevertheless, both fixed and optimal policies exhibit adaptive efforts to sustain the production level, with optimal policies demonstrating less uncertainties. Moreover, optimal policies consistently outperform the fixed policy, notably demonstrating higher average total rewards. This suggests the superior effectiveness of optimal policies in adapting to temperature variations.

In Figure~\ref{fig:TNW}, it is evident that different agricultural policies lead to varying management practices, affecting N fertilization and irrigation strategies. Comparatively, the fixed policy results in generally higher N and water inputs than optimal policies. On average, the fixed policy entails 149\% higher N usage and 341\% higher water usage than optimal policies. Consequently, optimal policies achieve significantly higher rewards, although the resulting yields are slightly higher than the fixed policy.

Interestingly, optimal policies exhibit a substantial reduction in nitrate leaching compared to the fixed policy, but they result in higher N2O emissions, as illustrated in Figure~\ref{fig:TNL}. This unexpected outcome contradicts our initial expectations and can be partially explained by the reward function defined in Eqn.~(\ref{eq:AgReward}). In the pursuit of maximizing the total reward, the agent seeks a delicate balance between gains, such as corn yield, and penalties, encompassing fertilizer and water usage, nitrate leaching, and N2O emission. It becomes apparent that, in the process of learning optimal policies, the agent prioritizes the optimization of N and water usage and nitrate leaching mitigation at the expense of minimizing N2O emissions to enhance the overall reward. Further discussions and alternative solutions are presented in the conclusion section.  

\subsection{Precipitation reducing}

We also investigate the impact of reduced rainfall on fertilization and irrigation management, as well as agricultural outcomes. After analyzing historical rainfall data dating back to 1950, we identified no consistent trend in annual rainfall. In our study, we base our simulations on the actual weather conditions from 2012 but make adjustments by decreasing the monthly average rainfall by 20\%, 40\%, 60\%, and 80\%, respectively, throughout the year while keeping the monthly maximum and minimum temperatures consistent, mirroring those of 2012. It's important to note that scenarios involving increased precipitation levels that may result in flood-related crop damage are not considered, as such situations fall beyond the forecasting capabilities of DSSAT.

\begin{figure}
\centering
\resizebox*{10cm}{!}{\includegraphics{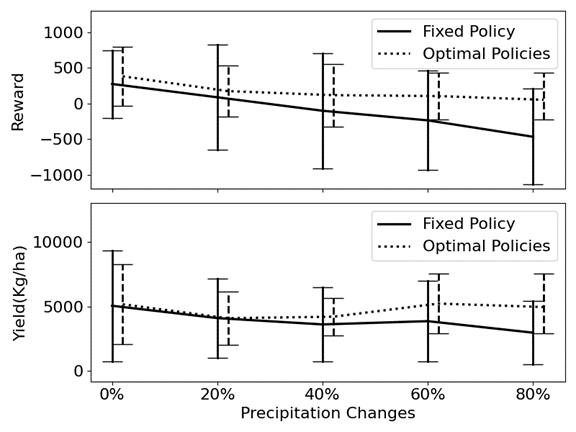}}
\caption{Reward and yield with 95\% PI from different policies when monthly precipitation reduces.} \label{fig:PRY}
\end{figure}

\begin{figure}
\centering
\resizebox*{10cm}{!}{\includegraphics{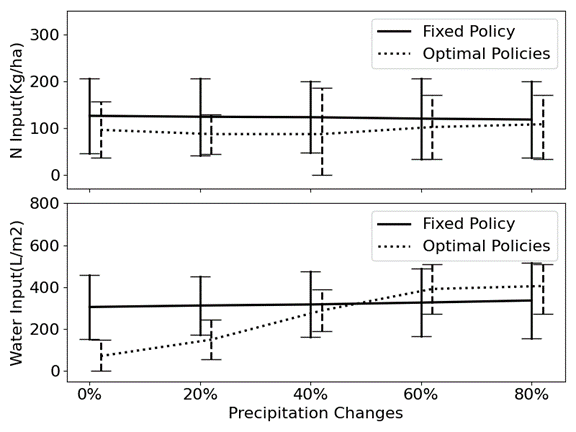}}
\caption{N and water input with 95\% PI from different policies when monthly precipitation reduces.} \label{fig:PNW}
\end{figure}

\begin{figure}
\centering
\resizebox*{10cm}{!}{\includegraphics{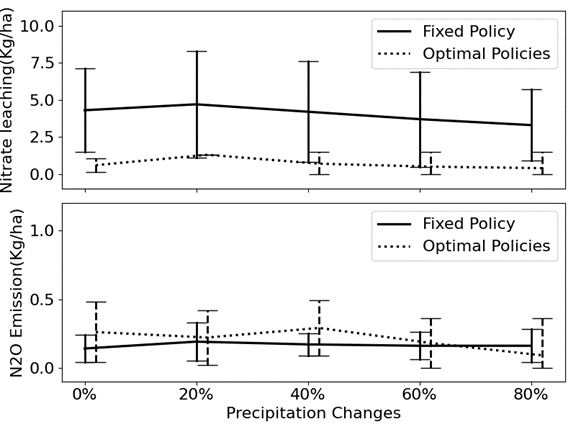}}
\caption{Nitrate leaching and N2O emission with 95\% PI from different policies when monthly precipitation reduces.} \label{fig:PNL}
\end{figure}

Aligned with our findings in the study of temperature variability, Figure~\ref{fig:PRY} illustrates that optimal policies also exhibit superior performance compared to the fixed policy in scenarios of reduced precipitation. Optimal policies result in larger harvests and rewards, particularly under more severe conditions, such as an 80\% reduction in precipitation, representing drought events. In these instances, optimal policies achieve an average yield increase of 120\% and demonstrate enhanced efficiency.

Figure~\ref{fig:PNW} provides insights into the factors contributing to this outcome by comparing N and water usage between the fixed and optimal policies. The fixed policy exhibits limited responsiveness to precipitation reduction, maintaining constant N and water usage. Although N inputs remain relatively stable, optimal policies display sensitivity to reduced rainfall by adjusting water input accordingly. In the case of a severe drought event with an 80\% short of rainfall, the average water input increases by 300\% to sustain the same corn yield. In response to precipitation reduction, optimal policies demonstrate greater adaptability to climate variability. 

\section{Conclusion and future works}

Addressing global hunger and lessening environmental consequences requires a careful balance between maximizing crop yield and limiting GHG emissions from agricultural activities. This study marks the first attempt to integrate considerations of N2O emissions into the optimization of agricultural management, with a particular focus on adapting to climate variability. Using a model-free RL method, specifically DQN with RNN-based Q networks, our research aims to train intelligent agents that learn optimal management strategies or policies to efficiently handle N fertilization and irrigation, ultimately reducing N2O emissions, minimizing nitrate leaching, and maximizing crop yields.  

In this study, we account for two significant sources of uncertainty. First, a probabilistic ML model is developed to estimate N2O emissions throughout the crop growth phase. This model, which adopts the MaxLike approach to address data uncertainty, enhances the capabilities of the deterministic model. The incorporation of this probabilistic element contributes to a more comprehensive and insightful prediction framework. Secondly, to introduce variability in weather conditions, a stochastic weather generator, WGEN, is integrated into the crop simulator (Gym-DSSAT). WGEN generates random weather scenarios based on actual weather data, further enriching the study's exploration of the agent's resilience to climate change. 

The results indicate that, by penalizing N2O emissions in the reward function, the agent can successfully balance crop yield, N and water usage, nitrate leaching, and N2O emissions, providing optimal policies. Our research extends the application of the developed framework to assess the impact of climate variability on agricultural results and practices. Specifically, we focus on scenarios involving elevated temperatures and limited rainfall. The findings reveal that the previously established policy is resilient to variations in temperature and mild changes in precipitation, but it faces challenges under severe conditions, such as extremely substantial reductions in rainfall or droughts. In contrast, the optimal policies learned based on specific weather conditions are more adaptive, particularly in light of extreme climatic events. 

The simulations in this study rely exclusively on data from 2012, particularly the daily N2O emission data. Feature selection for the ML models to estimate N2O emission aligned with observable state variables, potentially excluding important features and limiting model performance. Notably, the KBS-LTER testing field experienced a one-time fertilization with high N input, differing from the agent's preferred multiple-and-small fertilization strategy it learns. Future plans include compiling an extensive dataset of historical daily GHG emissions under various N input scenarios specific to the relevant agricultural areas. By incorporating this comprehensive dataset, we aim to conduct simulations that are both more representative and accurate, considering a broader range of soil-derived GHGs such as N2O, NOx, and others. This approach will significantly enhance the trustworthiness and applicability of our results.

In this research, we tackle N2O emission by introducing an additional term in the reward function. The results depict that the agent may prioritize maximizing crop yield at the potential expense of minimizing the N2O emission in order to achieve the highest total reward. As we look ahead, we plan to explore Multi-Objective Reinforcement Learning (MORL) as a potential solution that can potentially enable the simultaneous optimization of multiple conflicting objectives \cite{Hayes2022}, such as maximizing crop yield while minimizing N2O emissions. By employing MORL, we will create a more nuanced reward structure that better reflects the complexity of agricultural decision-making, ensuring that environmental considerations are weighed alongside economic ones.

Another alternative under consideration involves leveraging formal logic language to express the N2O budget as a specification. This specification can then be transformed into a finite state automaton and seamlessly integrated into the RL framework \cite{Li2023}. By adopting this approach, the N2O budget can be enforced through model-checking techniques. These potential approaches aim to enhance the agent's decision-making capabilities regarding crop yield and N2O emission in a more nuanced and optimized manner.

Furthermore, our future endeavors include gathering comprehensive cost data for the relevant year, encompassing expenses such as fertilizer, water, machinery, labor, and other operational costs. Additionally, we plan to integrate economic elements such as agricultural subsidies offered by the government and possible inflation in the upcoming years. Incorporating these financial factors into our model will enable it to more accurately reflect farmers' net income. This enhancement will significantly elevate the contribution and impact of our model, offering a more holistic understanding of the economic implications of the optimized agricultural strategies proposed.

\section*{CRediT authorship contribution statement}
Zhaoan Wang: Conceptualization; Methodology; Software; Validation; Investigation; Data Curation; Writing - Original Draft; Writing - Review \& Editing; Visualization.

Shaoping Xiao: Conceptualization; Methodology; Investigation; Writing - Review \& Editing; Supervision; Project administration; Funding acquisition.

Jun Wang: Conceptualization; Writing - Review \& Editing; Supervision; Project administration; Funding acquisition.

Ashwin Parab: Software; Validation.

Shivam Patel: Software; Validation.

\section*{Declaration of Competing Interest}
The authors declare that they have no known competing financial interests or personal relationships that could have appeared to influence the work reported in this paper.

\section*{Acknowledgments}
Wang Z, Xiao S, and Wang J received support from the University of Iowa OVPR Interdisciplinary Scholars Program for this study. Wang Z and Xiao S also received support from the U.S. Department of Education (E.D. \#P116S210005).

\section*{Declaration of Generative AI and AI-assisted technologies in the writing process}
During the preparation of this work, the authors used GPT-3.5 and GPT-4 to improve the readability and language during the writing process. After using this tool/service, the authors reviewed and edited the content as needed and took full responsibility for the content of the publication.

\end{document}